\documentclass[letterpaper,10pt,conference]{IEEEtran}

\usepackage{copyright}

\IEEEoverridecommandlockouts

\usepackage{graphicx}
\usepackage{booktabs}

\usepackage{amsmath}
\usepackage{amssymb}
\usepackage{mathtools}
\usepackage{physics}
\usepackage{xfrac}

\usepackage{pifont}
\newcommand{\cmark}{\ding{51}}%
\newcommand{\xmark}{\ding{55}}%

\usepackage{threeparttable}
\usepackage{subcaption}
\usepackage{tabularx}

\DeclarePairedDelimiterX\set[1]\lbrace\rbrace{\def\given{\;\delimsize\vert\;}#1}

\usepackage{cleveref}
\crefname{table}{Tab.}{Tabs.}
\crefname{figure}{Fig.}{Figs.}
\crefname{section}{Sec.}{Secs.}
\crefname{equation}{Eq.}{Eqs.}

\usepackage[acronym]{glossaries}
\newacronym{auc}{AUC}{area-under-curve}
\newacronym[longplural=Degrees-of-Freedom]{dof}{DoF}{Degree-of-Freedom}
\newacronym{svd}{SVD}{Singular Value Decomposition}
\newacronym{ransac}{RANSAC}{RANdom SAmple Consensus}
\newacronym{dbscan}{DBSCAN}{Density-Based Spatial Clustering of Applications with Noise}
\newacronym{mlp}{MLP}{Multi-Layer Perceptron}
\newacronym{rmse}{RMSE}{Root Mean Square Error}
\newacronym{icp}{ICP}{Iterative Closest Point}
\newacronym{fpfh}{FPFH}{Fast Point Feature Histograms}
\newacronym{fcgf}{FCGF}{Fully Convolutional Geometric Features}
\newacronym{cnn}{CNN}{Convolutional Neural Network}
\newacronym{miou}{mIoU}{mean Intersection over Union}
\newacronym{rte}{RTE}{Relative Translation Error}
\newacronym{rre}{RRE}{Relative Rotation Error}

\usepackage{cuted}

\usepackage{tikz}
\usetikzlibrary{fit,calc}

\usepackage{float}
\usepackage{nidanfloat} 

\usepackage{pgf} 
\usepackage{pgfplots}
\usepgfplotslibrary{external}
\usepackage{pgfplotstable}
\usepackage{pgfmath}
\usepgfplotslibrary{statistics}
\pgfplotsset{compat=1.8}

\usepackage{filecontents}

\usepackage[binary-units]{siunitx}

\usepackage{scalefnt}

\usepackage{xcolor}
\newcommand{\gp}[1] {{\color{red} gp: #1}}

\usepackage{paralist}

\begin{document}

%------------------------------------------------------------------
\title{
\Large
\bf
% Semantic Graph Descriptors for Simultaneous Place Recognition and Pose Estimation from LiDAR
BoxGraph: Semantic Place Recognition and Pose Estimation from 3D LiDAR
}
\author{Georgi Pramatarov, Daniele De Martini$^{*}$, Matthew Gadd$^{*}$, and Paul Newman
\\
Mobile Robotics Group (MRG), University of Oxford, $^{*}$\textit{Equal contribution}\\\texttt{\{georgi,daniele,mattgadd,pnewman\}@robots.ox.ac.uk}
}
\maketitle
%------------------------------------------------------------------

% notice
\copyrightnotice

%------------------------------------------------------------------
\begin{abstract}
This paper is about extremely robust and lightweight localisation using LiDAR point clouds based on instance segmentation and graph matching.
% For both metric pose estimation and place recognition we model 3D point clouds as fully-connected graphs of semantically identified components where each vertex in the representation corresponds to an object instance and is encoded by considering its shape.
We model 3D point clouds as fully-connected graphs of semantically identified components where each vertex corresponds to an object instance and encodes its shape.
Optimal vertex association across graphs allows for full 6-\gls{dof} pose estimation and place recognition by measuring similarity.
This representation is very concise, condensing the size of maps by a factor of $25$ against the state-of-the-art, requiring only \SI{3}{\kilo\byte} to represent a \SI{1.4}{\mega\byte} laser scan.
We verify the efficacy of our system on the SemanticKITTI dataset, where we achieve a new state-of-the-art in place recognition, with an average of \SI{88.4}{\percent} recall at \SI{100}{\percent} precision where the next closest competitor follows with \SI{64.9}{\percent}.
We also show accurate metric pose estimation performance -- estimating 6-\gls{dof} pose with median errors of \SI{10}{\centi\metre} and \SI{0.33}{\deg}.
% of the true vehicle position.
% This allows for full 6-\gls{dof} pose estimation as well as graph matching to recognise similar places
% while condensing the size of the map by a factor of $25$ with respect to the state-of-the-art, requiring only approximately \SI{3}{\kilo\byte} to represent a \SI{1.4}{\mega\byte} scan for matching.
% On top of these run-time performance gains, we verify the efficacy of our system on several sequences of the SemanticKITTI dataset showing competitive localisation performance -- primarily achieving \SIrange{80}{100}{\percent} recall at \SI{100}{\percent} precision on average in place recognition while estimating metric pose to within \SIrange{0}{15}{\centi\metre} of the true vehicle position.
\end{abstract}
% \begin{IEEEkeywords}
% Localisation, Place Recognition, Pose Estimation, Semantic Segmentation, Semantic Mapping, Autonomous Vehicles, Robotics
% \end{IEEEkeywords}

% \pmnsays{I like these words. But might we not empahsis here that the reduction of maps size for the dual task we might use a map for (localisation and seeding) is a core purpose. Also I think it is stronger to say the system has all the components tyou describe by design, so maybe not "here is system A and if we do this extra thing it gets better" }

\glsresetall

%------------------------------------------------------------------
\section{Introduction}%
\label{sec:introduction}
%------------------------------------------------------------------

% \begin{strip}
\begin{figure*}[b]
\centering
\includegraphics[width=0.85\textwidth]{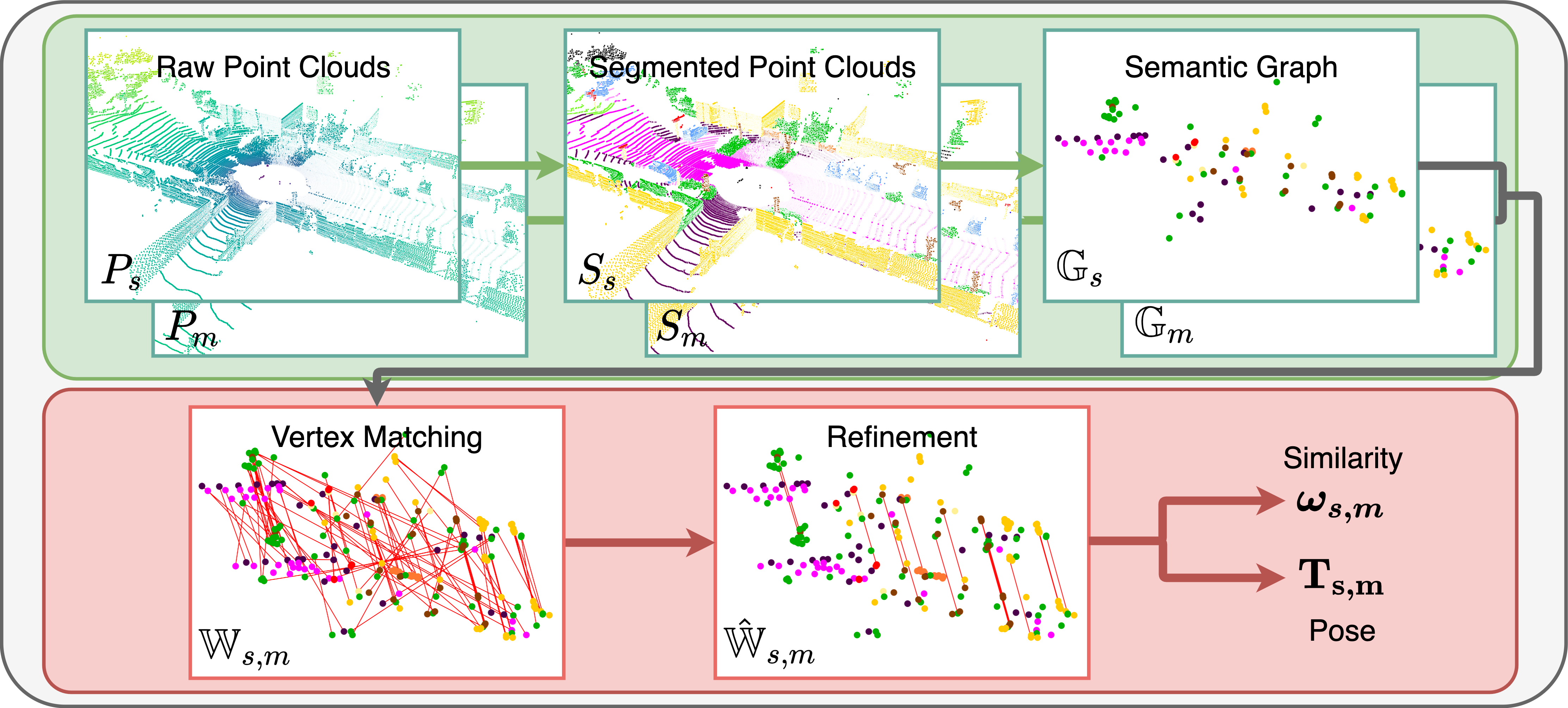}
\captionof{figure}{System diagram of our method -- raw point clouds, $P_s$ and $P_m$, are labelled by a learned segmentation model and then clustered based on class to obtain compact complete graphs of semantic vertex instances, $\mathbb{G}_s$ and $\mathbb{G}_m$.
The instances are then matched according to their shape similarity, and the resulting correspondences $\mathbb{W}_{s, m}$ are refined to produce the final similarity score and pose estimate between the two scenes.
\label{fig:method_diagram}}
% \end{strip}
\end{figure*}

The ability to localise in an already visited location is crucial for autonomous operations.
A commonly used approach is to build a map with the same sensor used during operations.
Still, appearance change -- due to change of viewpoint, occlusions caused by dynamic objects, or varying weather conditions -- strongly affects the performances of such systems.
LiDAR sensors find themselves to be a very popular choice for autonomous vehicles due to their superior robustness to these issues.
Indeed, the fact that they actively sense the environment makes them less affected by appearance change and lighting conditions, and the geometric nature of their readings contains rich information about their surroundings.

Classically, LiDAR-based loop-closure detection methods rely on either local keypoints or global features.
Unfortunately, the first emphasise local details -- and result in huge maps -- while the second discard much of the useful properties contained in the geometric distribution of the point cloud, often making metric estimation infeasible from the same data structures.
Recently, semantic-driven loop-closure detection has shown that, by defining object-wise features, it is possible to produce \textit{reliable and condensed} maps that contain both high-level information about the objects in the scene and their spatial relationships.
A new vision-based approach~\cite{qin2021semantic} models input images of an indoor scene as fully-connected graphs of objects.
% and phrases loop-closure detection as a graph-matching problem.
These objects are detected by a network and described by the dimensions of their enclosing bounding boxes.
Graph-matching and similarity measurement then amounts to finding optimal correspondences between objects.
% in two images.

% uses a detection module to extract bounding boxes of the objects in the scene and builds a fully-connected graph model

%\gp{Mention that we extend the camera based method\cite{qin2021semantic}? Or discuss in the relevant work section and just reference later every time we do something similar}
Motivated by the success of this approach, we propose a global localisation method for 3D LiDAR point clouds that represents scenes as fully-connected semantic graphs~(\cref{fig:method_diagram}).
% Information on objects' type and appearance is stored in the vertices and relative object positions in the environment in the edges.
Information on objects' type and appearance is stored in the vertices while their relative positions in the environment are stored in the edges.
In doing so, we capture both local and global geometrical features and higher-level semantic properties of a scene, resulting in powerful expressiveness.
The place recognition task then reduces to an optimal assignment problem between the vertices of two graphs.
In contrast to~\cite{qin2021semantic}, we develop this method further and show that it can yield a 6-\gls{dof} metric pose estimate.

Our principal contributions are (1) the specialisation of a camera-based semantic graph-based place recognition system to 3D point clouds (using semantic blobs in 3D instead of 2D bounding boxes) and (2) the extension of this to allow for precise pose estimation on top of place recognition.
% \begin{enumerate}
% \item We specialise a camera-based semantic graph-based place recognition system to 3D point clouds,
% \item We extend this to allow for precise pose estimation on top of place recognition,
% \item We demonstrate the viability of the system for outdoor LiDAR-based place recognition and extend it to 6-\gls{dof} pose estimation.
% which models a scene as a concise fully-connected graph of semantic entities. The entities are extracted from raw data using a laser segmentation network and a Euclidean clustering procedure as opposed to a 2D object detection algorithm.
% \item We introduce a novel application of a simple contrastive learning framework utilised for feature extraction from vertex entities that improves upon the hand-crafted variant in both pose estimation and longer-range place recognition.
% \item A semantic graph description of the world as scanned by LiDAR via RangeNet++ segmentation and DBSCAN clustering
% \item Features extracted from the graph itself suitable for both place recognition and rigid-body pose estimation
% \item Improved vertex descriptors by incorporating some appearance alongside semantic properties, leading to better long-range performance at both place recognition and rigid-body pose estimation
% \end{enumerate}
We validate the approach on the KITTI odometry dataset and achieve against contemporary methods a \textit{new state-of-the-art} in terms of place recognition (as compared to all approaches) and compactness (as compared to methods capable of both place recognition and metric pose estimation).
We also achieve precise semantic-based pose estimation, where our performance is competitive with geometry-based approaches that operate on dense point clouds.
Indeed, our evidence suggests that if given increasingly good segmentation predictions (which are not the focus of this paper), our method may challenge geometry-based methods in pose-estimation precision.
% \gp{Maybe go stronger on the metric results}
% \gp{
% [13]: \cite{qin2021semantic}
% sensor modality, indoor-outdoor, object detection-semantic segmentation, ransac refinement, don't do metric pose est
% }
% \gp{
% gos-match \cite{zhu2020gosmatch}
% vertix desc are histograms, don't consider appearance/actual objects,
% similarity edge driven not vertex driven,
% they use ransac to verify, we use ransac before to improve matches
% }

% We proceed by reviewing relevant literature in~\cref{sec:related} before detailing the method in~\cref{sec:method}.
% \Cref{sec:experiments} presents the experimental setup used to validate the theoretical contribution of this work, with subsequent results and conclusions covered in~\cref{sec:results} and~\cref{sec:conclusion} respectively.

%------------------------------------------------------------------
\section{Related Work}%
\label{sec:related}
%------------------------------------------------------------------

Traditional localisation techniques on 3D point clouds typically exploit low-level geometrical information. Approaches based on matching local keypoints \cite{rusu2009fpfh,kallasi2016fastkeypoint,choy2019fully} can yield precise pose estimates but rely on storing dense data and do not translate to place recognition.
Matching of geometric primitives such as lines can be susceptible to occlusions~\cite{suleymanov2020lidar}.
Instead, global descriptor-based approaches abstract coarse information, becoming more susceptible to scene ambiguity, and are rarely coupled with accurate registration capabilities.
For example, M2DP~\cite{he2016m2dp} projects a point cloud to several 2D planes and extracts density signatures for each plane, which could also be vulnerable to occlusions.
Scan Context-based methods~\cite{kim2018scan, wang2020isc} reduce a 3D point cloud to a flat polar grid, which could be affected by translational offsets.
LiDAR-Iris~\cite{wang2021iris} produces binary signature images from bird's-eye-view projections of point clouds, also encoding the vertical distribution of points, which can suffer under roll and pitch variation.
LocNet~\cite{yin2020locnet} proposes a histogram-based descriptor, contrastively tuned given similar scenes. However, this is still a much denser representation of the environment than our descriptor and only estimates a 3-\gls{dof} pose.

The advances of point-cloud learning methods such as PointNet~\cite{qi2017pointnet} and PointNet++~\cite{qi2017pointnetpp} have shown their applicability in the domain of localisation: PointNetVLAD~\cite{uy2018pointnetvlad} and LPD-Net~\cite{liu2019lpdnet} combine learning approaches with the NetVLAD descriptor~\cite{arandjelovic2016netvlad} to obtain a global scene representation; however, they discard information about the distribution of objects, and so they are not directly applicable to pose estimation.
Segment-based methods such as SegMatch~\cite{dube2017segmatch} and SegMap~\cite{segmap2019dube} partition point clouds into higher-level segments and, similarly to our approach, extract features for each segment and match them across scenes.
In contrast, however, they do not consider the relative positions between segments in each scan.
Locus \cite{vidanapathirana2020locus}, also a segment-based method, extracts a global descriptor from topological and temporal segment features, which depends on sequential data aggregation, thus requiring an autonomous system to move before initialising.

To improve the representational power of 3D descriptors, various methods employ  semantics as additional inputs.
Semantic Scan Context~\cite{li2021ssc}, for instance, extends~\cite{kim2018scan} by encoding semantic information into the polar grid, demonstrating promising results; still, it only produces a rough 3-\gls{dof} pose estimate.
OverlapNet~\cite{chen2021overlapnet} combines semantic, geometric, and intensity information and uses a \gls{cnn} to estimate the relative yaw and overlap between laser scans.
Instead, we explicitly extract information about the instances in a scene to create a higher-level -- and compact -- semantic and geometric representation.

SemSegMap~\cite{cramariuc2021semsegmap}, a semantic extension to~\cite{segmap2019dube}, projects RGB and semantic information from an input camera feed to point clouds before building segments that show increased stability.
They employ a contrastive learning scheme for segment feature extraction by self-supervising with segments from a pre-built map; however, this again relies on the aggregation of point clouds.
PSE-Match~\cite{yin2021pse} partitions point clouds per semantic class, extracts viewpoint-invariant spherical representations, and combines them into global place descriptors.
They incorporate inter-class divergence learning at semantic instead of instance level, thus disregarding underlying relationships between objects of the same class.

Recently, graph-based approaches have shown potential in the global localisation domain when combined with semantic information.
Both~\cite{kong2020semantic}~and~\cite{zhu2020gosmatch}, similarly to us, create complete graphs of clustered instances from semantically-segmented point clouds.
In~\cite{kong2020semantic}, a neural network architecture is applied to extract vertex features based on graph connectivity and then predict the similarity of two graphs for place recognition.
Thus, the system requires paired scans for supervision, while our approach can use pure geometric vertex features without training.
GOSMatch~\cite{zhu2020gosmatch}, instead, computes histogram-based descriptors from the distances between the objects in the scene -- a global graph descriptor for loop closure detection and vertex descriptors for initial pose estimation and verification.
% -- from a semantically segmented point cloud where only cars, poles and trucks are considered as vertices.
These approaches, however, consider only the spatial arrangement of object instances -- via the graph edges.
This motivates us to model both the relative distances and the individual objects' shapes.
We show that even a simple object shape descriptor such as its bounding box can produce significant results in global localisation.

\begin{figure}[!h]
\centering
\includegraphics[width=0.78\columnwidth,trim={4.2cm 4.7cm 3.7cm 4.5cm},clip]{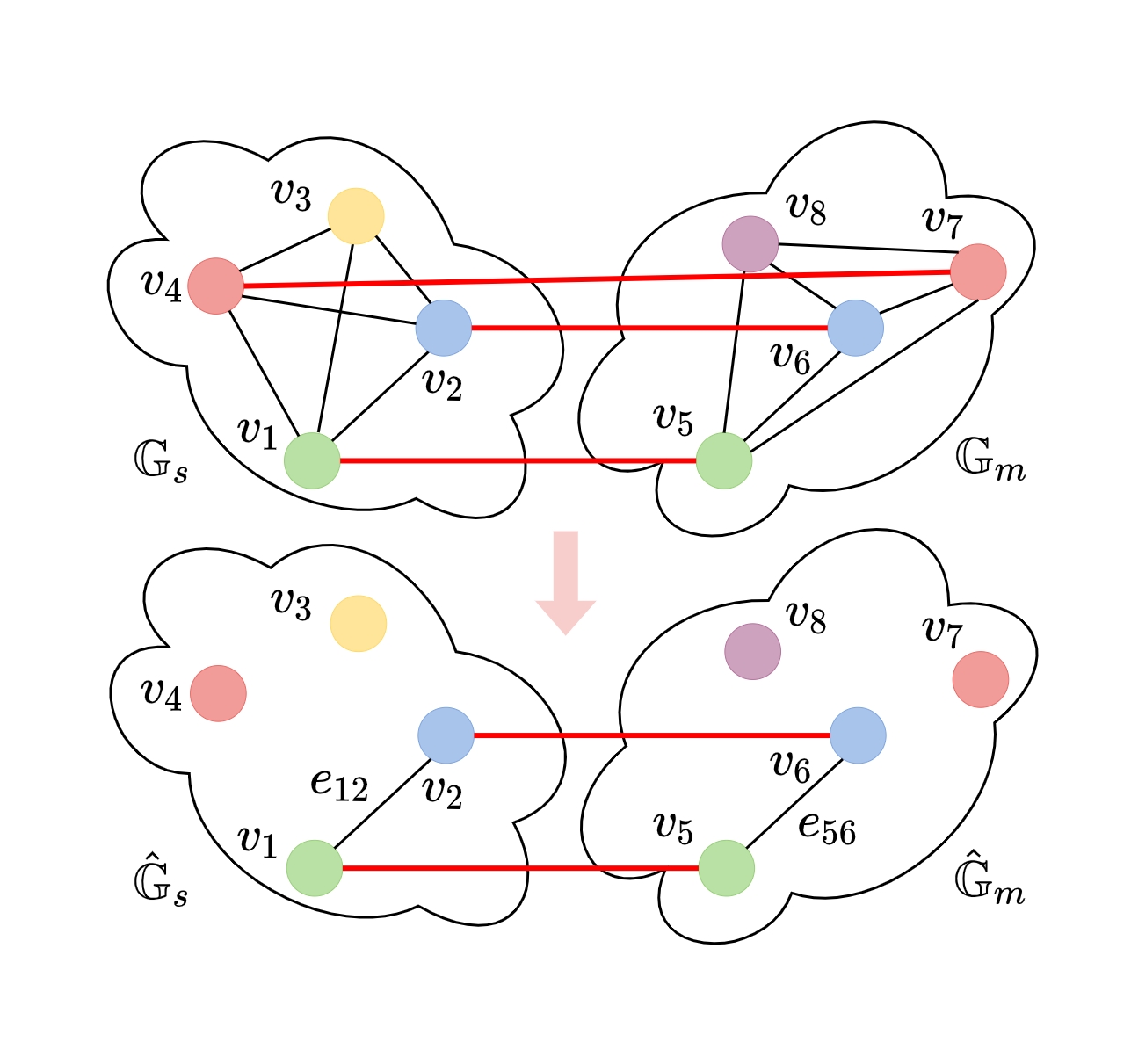}
\caption{
% \gadd{bigger symbols, bolder colours, bolder lines, etc}
Descriptor-matching amounts to finding the optimal assignment between same-class vertices based on shape similarity $\sigma_v(v_i)$.
RANSAC removes outliers (e.g. $v_4$ to $v_7$ above) and results in a relative transformation $T_{s,m}$.
The similarity score takes into account the similarity between matching vertices and edges of the refined graphs $\hat{\mathbb{G}}_s$ and $\hat{\mathbb{G}}_m$: $\omega_{s,m} = \sigma_e(e_{12}, e_{56}) + \sigma_v(v_1, v_5) + \sigma_v(v_2, v_6)$.
\label{fig:graph_matching}}
\vspace{-17pt}
\end{figure}

\section{Method}%
\label{sec:method}
%------------------------------------------------------------------

Our approach to semantic localisation in the LiDAR domain is depicted in~\cref{fig:method_diagram}.
We frame our system as a topometric localisation problem, where a point cloud
\begin{equation}
P_s: \set{p_1 \ldots p_n \given p_i \in \mathbb{R}^3 }
\end{equation}
from a LiDAR sensor stream $\mathbb{P}$ is matched against point clouds $P_m$ belonging to a pre-built map $\mathbb{M}$.

We approach this problem by exploiting the semantic information contained in $P_s$ and $P_m$ and represent them as fully-connected graphs, $\mathbb{G}_s$ and $\mathbb{G}_m$, to model objects -- with their type and shape -- and their relative positions.
Each object is translated in a vertex that encodes its type, appearance and spatial location.

We then solve the localisation problem by finding the optimal association between the vertices of two such graphs that maximises appearance similarity.
This is used to produce both a 6-\gls{dof} pose estimate $T_{s,m} = [R_{s,m} | t_{s,m}]$ -- where $R_{s,m}$ and $t_{s,m}$ are respectively the rotation matrix and the translation vector -- and a final similarity score $\omega_{s, m}$ useful for place recognition.

\subsection{Semantic Graph Extraction}%
To build a meaningful graph representation of a scene for localisation, distinctive semantic information needs to be extracted. 
This means a natural preference for stable objects -- e.g.~buildings and vegetation -- over ephemeral and dynamic ones -- such as vehicles and pedestrians -- which we exclude.
In contrast to~\cite{qin2021semantic} who use indoor object detection, we use semantically segmented point clouds as input to our system, since outdoor object detection and instance segmentation traditionally only focuses on dynamic objects.

Let $\mathbb{L}$ be the set of the semantic classes of interest and define $\lambda: \mathbb{R}^3 \rightarrow \mathbb{N}$ to be a label function that can assign a semantic label $l \in \mathbb{L}$ to every point $p \in P$, resulting in a semantically enhanced point cloud
\begin{equation}
S = \set{s_i \given s_i = \{p_i, \lambda(p_i)\} ~\forall p_i \in P }
\end{equation}
where each $s_i \in S$ contains the semantic label $l_i$ alongside the three-dimensional position of the corresponding point $p_i$.

We apply the \gls{dbscan} algorithm~\cite{ester1996density} to cluster points with the same label based on their Euclidean distance.
This results in a set of clusters of points, each constituting a point cloud ``local'' to a region of the scan that shares the same semantic class
\begin{equation}
\begin{split}
\mathbb{O} = \{ O_1, \ldots, O_M~\vert~&O_k \subset P,\\&l_k=\lambda(p_i) =\lambda(p_j) ~\forall p_i, p_j \in O_k \}
\end{split}
\end{equation}

Each cluster in $\mathbb{O}$ is then used as a vertex to produce a fully-connected, semantic graph descriptor $\mathbb{G} = \langle \mathbb{V}, \mathbb{E}\rangle$.
A vertex $v_i = \langle c_i, l_i, f_i \rangle \in \mathbb{V}$ is described by the geometrical centroid $c_i \in \mathbb{R}^3$ and the semantic class $l_i$ of the corresponding cluster $O_i$, and a feature vector $f_i \in \mathbb{R}^D$ extracted from the cluster's points and providing a compact representation of an object's appearance, as discussed further in~\cref{sec:method:featureextraction}.
An edge $e_{ij} = \langle v_i, v_j\rangle \in \mathbb{E}$, instead, is described by the two centroids $c_i$ and $c_j$ of the vertices $v_i$ and $v_j$ respectively.
In this view, the vertices model an approximation of the objects that compose the scene, whereas the edges model their relative position.

\subsection{Vertex Feature Extraction and Vertex Similarity}
\label{sec:method:featureextraction}
To compare scenes and their associated semantic graphs, we need to extract objects' appearances and measure their similarity.
% To do so, we consider two alternatives for extracting a feature vector $f_i$ from a cluster $O_i$: 
% \begin{inparaenum}[(1)]
%     \item through bounding boxes, and
%     \item through a learned approach.
% \end{inparaenum}
We use the feature vector of two vertices to compute the similarity between them, defined as $\sigma_v: \mathbb{V}\times\mathbb{V}\rightarrow\mathbb{R}$.
% Both feature extraction and corresponding vertex similarity are described in the following.

% \subsubsection{Bounding box shape features}
In contrast to \cite{qin2021semantic}, we start with 3D semantic blobs (not 2D detected image patches) and extract descriptors of the properties of scene entities.
Bounding boxes provide a rough yet compact approximation of a cluster's appearance: intuitively, graphs of nearby scans result in similar clusters.
In addition, the heading of a vehicle is typically aligned with the road in urban settings, so it is reasonable to assume that clusters' bounding boxes will remain consistent across revisits.
For this reason, we calculate the bounding box of each cluster and use its dimensions as the feature vector, i.e. $f_i = (h_i, w_i, d_i) \in \mathbb{R}^3$, where $h_i$, $w_i$, and $d_i$ are the elongation of the cluster in the $z$, $y$ and $x$ directions respectively, measured along the LiDAR frame of reference mounted on the vehicle.
If we let
\begin{equation}
\delta(a, b) = \frac{|a - b|}{\max{(a, b)}}
\end{equation}
% the similarity function $\sigma_v^b$ can be defined as:
then a vertex similarity function $\sigma_v$ can be defined as:
\begin{equation}
\small{
% \sigma_v^b(v_i, v_j) = \exp(-\frac{\delta(h_i, h_j) + \delta(w_i, w_j) + \delta(d_i, d_j)}{3})
\sigma_v(v_i, v_j) = 
\begin{cases}
      \exp(-\frac{\delta(h_i, h_j) + \delta(w_i, w_j) + \delta(d_i, d_j)}{3})&\!\!\!\!\!\text{if}\ l_i=l_j \\
      0, & \!\!\!\!\!\text{otherwise}
    \end{cases}
    }
\end{equation}
Here $\delta$ computes the fractional variation of each dimension of the bounding boxes, and the average of such a discrepancy is used as the shape similarity metric.

\subsection{Graph Matching}
 
Let $\mathbb{G}_s = \langle \mathbb{V}_s, \mathbb{E}_s \rangle$ and $\mathbb{G}_m = \langle \mathbb{V}_m, \mathbb{E}_m \rangle$ be the graph representation of the point clouds $P_s \in \mathbb{P}$ and $P_m \in \mathbb{M}$.
To calculate the similarity of the two semantic graphs, we need to find the optimal correspondence between their vertices.
Let us assume for simplicity that $|\mathbb{V}_s| \leq |\mathbb{V}_m|$ without loss of generality.
We seek to find a set of matches $\mathbb{W}_{s, m} = \set{ (v_i, v_j) \in \mathbb{V}_s \times \mathbb{V}_m}$, where each $v_i$ is matched to one and only one $v_j$, such that $\sum_{\mathbb{W}_{s, m}} \sigma_v(v_i, v_j)$ is maximal.
% \gp{add big sigma for sum, or generally define better} Note that $|MATCHES| = \min{(|V|, |V'|)}$, so there exist unmatched vertices in the larger graph, but each vertex in the smaller one is matched.
This is a classical instance of a linear assignment problem, which can be solved in polynomial time using the Hungarian algorithm, similarly to~\cite{qin2021semantic}.
%\gp{More details about the Hungarian algorithm? - The problem setup for the Hungarian algorithm is, given a weighted bipartite graph, find a matching that minimises the sum of edge weights. In our setting, a we define an undirected bipartite graph $\mathcal{G}\ = (\mathcal{V}, \mathcal{E})$ where $\mathcal{V} = V \cup V'$ and $\mathcal{E} = \set{(v, v') \in V \times V'}$ and edge weights equal to the similarity score between two vertices. That is, each vertex in $V$ is connected to each vertex in $V'$. We want to find a maximal matching, which is analogous. }
Moreover, as in \cite{qin2021semantic}, we improve the computational cost of this procedure by matching only the vertices of the same class, i.e. with non-zero similarity~(\cref{fig:graph_matching}).

\subsection{Pose Estimation and Refinement}
\label{sec:poserefine}

After obtaining the optimal vertex correspondences $\mathbb{W}_{s, m}$ between two graphs, in contrast with~\cite{qin2021semantic}, we apply the \gls{svd} technique on vertex centroids to estimate a 6-\gls{dof} pose.
As discussed above, however, the matching procedure associates each vertex of the smaller graph with one from the larger graph.
Since this could lead to incorrect matches reducing the quality of the resulting pose estimate, we employ the \gls{ransac}~\cite{fischler1981ransac} algorithm to discard the outliers: we apply \gls{svd} iteratively on a subset of the matches to find the most inliers, subject to a tolerance threshold.
The resulting transform represents the final pose estimate $T_{s,m} \in SE(3)$.
In addition, the inliers form a refined set of matches $\hat{\mathbb{W}}_{s, m} \subseteq \mathbb{W}_{s, m}$, which we use next to calculate the graph similarity score~(\cref{fig:graph_matching}).
% This capability was not developed in~\cite{qin2021semantic} and is presented for the first time here.

\subsection{Similarity Score}%
\label{sec:simscore}
Let $\omega_{s, m}$ be the similarity score between $\mathbb{G}_s$ and $\mathbb{G}_m$.
It is designed to take into account both the vertex similarities from the refined set -- which include semantics and appearance information -- and the similarities of edges with matching endpoints -- which contain the geometrical relationships between the objects; \cref{fig:graph_matching} shows a stylised example.
In particular, the inlying set of matches $\hat{\mathbb{W}}_{s,m}$ is used to produce two fully-connected sub-graphs $\hat{\mathbb{G}}_s = \langle \hat{\mathbb{V}}_s, \hat{\mathbb{E}}_s\rangle \subseteq \mathbb{G}_s$ and $\hat{\mathbb{G}}_m = \langle\hat{\mathbb{V}}_m, \hat{\mathbb{E}}_m\rangle \subseteq \mathbb{G}_m$.
As discussed above, each $e_{ij} = \langle v_i, v_j\rangle \in \hat{\mathbb{E}}_s$ and $e_{lk} = \langle v_l, v_k\rangle \in \hat{\mathbb{E}}_m$ are described by the centroids of their corresponding vertices.
Similarly to \cite{qin2021semantic}, we define the edge similarity to be
\begin{equation}
\sigma_e(e_{il}, e_{jk}) = \exp (-\delta(\norm{c_i - c_l}_2, \norm{c_j - c_k}_2))
\end{equation}
where $e_{il} \in \hat{E}_s$, $e_{jk} \in \hat{E}_m$ and $\norm{\cdot}_2$ is the $l_2$-norm of a vector in $\mathbb{R}^N$.
The final graph similarity score is then the sum of edge similarities of edges with matching vertices and the vertex similarities in the refined set:

\begin{equation}
\begin{split}
\omega_{s,m} = \sum_{\hat{\mathbb{W}}_{s,m}} & \sum_{\hat{\mathbb{W}}_{s,m}} \sigma_e(e_{il}, e_{jk}) + \sum_{\hat{\mathbb{W}}_{s,m}} \sigma_v(v_i, v_j) \\ &\forall v_i, e_{il} \in \hat{\mathbb{G}}_s\ \text{and}\ \forall v_j, e_{jk} \in \hat{\mathbb{G}}_m
\end{split}
\end{equation}

We consider graphs to correspond to the same place if the similarity score between them is above some threshold $\tau \in \mathbb{R}$.

% on the corresponding sub-point cloud $P_c = \set{ p_1, \ldots, p_N_c \given p_i \in \mathbb{R}^3 }$

%------------------------------------------------------------------
\section{Experimental Results}%
\label{sec:experiments}
%------------------------------------------------------------------

% \gp{Add brief description of the setup after it's been finalised}

% \subsection{Dataset}

We evaluate the proposed approach on the KITTI odometry dataset~\cite{Geiger2012CVPR}.
The dataset includes \num{11} sequences containing laser data from a Velodyne HDL-64 sensor and ground truth poses.
For our experiments, we use the six sequences that feature loop closures -- \texttt{00}, \texttt{02}, \texttt{05}, \texttt{06}, \texttt{07}, and \texttt{08}.
In particular, \texttt{Sequence 08} has reverse loops and so is useful to assess the rotation invariance of our system.

We study our system's performance with ground-truth semantic annotations from SemanticKITTI~\cite{behley2019semantickitti} and predictions from trained networks.
In line with related work~\cite{li2021ssc,kong2020semantic,zhu2020gosmatch}, we train a RangeNet++ \cite{milioto2019rangenet} LiDAR segmentation model and provide an ablation analysis using predictions from a state-of-the-art network, Cylinder3D~\cite{zhou2020cylinder3d}.
Indeed, while the latter achieves a \gls{miou} of \num{67.8} on SemanticKITTI, the former only reaches \num{52.2} and thus is key to assess the robustness of our approach to prediction noise and allows for a fair comparison with contemporary methods.

\subsection{Map Compactness}

\begin{table}[t]
    \footnotesize
    \centering
    \renewcommand{\arraystretch}{1.2}
    \resizebox{\columnwidth}{!}{\begin{tabular}{lcccccc}
        \toprule
        & & \multicolumn{2}{c}{\textbf{Descriptor shape}} & \multicolumn{2}{c}{\textbf{Size (kB)}} \\
        \textbf{Method} & \textbf{MPE} & \textbf{Global} & \textbf{Vertex} & \textbf{Avg} & \textbf{Max} \\
        \midrule
        PointNetVLAD~\cite{uy2018pointnetvlad} & \xmark & $256$ & - & $1$ & $1$ \\
        M2DP~\cite{he2016m2dp} & \xmark & $192$ & - & $0.75$ & $0.75$ \\
        IRIS~\cite{wang2021iris} & \xmark & $80 \times 360\SI{}{\bit}$ & - & $3.52$ & $3.52$ \\
        ScanContext~\cite{kim2018scan} & (\cmark) & $20 \times 60$ & - & $4.69$ & $4.69$ \\
        SSC~\cite{li2021ssc} & (\cmark) & $50 \times 360$ & - & $70.3$ & $70.3$ \\
        OverlapNet~\cite{chen2021overlapnet} & (\cmark) & $360 \times 128$ & - & $180$ & $180$ \\
        SGPR~\cite{kong2020semantic} & \xmark & $100 \times 4$ & - & $1.56$ & $1.56$ \\
        GOSMatch~\cite{zhu2020gosmatch} & (\cmark) & $60 \times 6$ & $60 \times 3$ & $75.2$ & $168.75$ \\
        
        \midrule
        BG (Ours) & \cmark & - & $7$ & $2.87$ & $6.51$ \\
        \bottomrule
    \end{tabular}}
    \caption{Analysis of memory consumption of global localisation systems for storing a single scan using different place recognition methods. The baselines are categorised by whether they perform metric pose estimation (MPE), where (\cmark) denotes only a coarse estimate. Systems with \xmark~and (\cmark) would also store the raw point cloud (not added in the table, $1.41$ MB on average) and rely on other registration techniques to produce refined poses. Average and maximum number of vertices in a graph is taken to be $105$ and $238$. \label{tab:map_size}}
    \vspace{-17pt}
\end{table}

As described in~\cref{sec:method}, of \num{28} classes provided, we make use only of distinctive static objects (sidewalk, building, fence, vegetation, trunk, pole, and traffic-sign).
For example, we leave out the road class as it is empirically less informative and frequently occluded.
% In contrast with~\cite{zhu2020gosmatch} -- where the authors state that parked vehicles are quite distinctive -- we also leave out dynamic types.
Semantic graphs are then generated for each scan in the sequences and stored in map databases.

Throughout all the sequences, the average and maximum number of vertices in a graph are \num{105} and \num{238}, respectively, with a total dimension of vertex descriptors of \num{7} -- the centroid, the semantic label and the bounding box features.
\Cref{tab:map_size} includes the resulting dimensions of the graphs and compares the size of our representation with other methods.

As baselines, we select state-of-the-art place recognition methods based on geometric global descriptors such as 
PointNetVLAD~\cite{uy2018pointnetvlad}, M2DP~\cite{he2016m2dp}, LiDAR-Iris~(IRIS)~\cite{wang2021iris}, and Scan Context~\cite{kim2018scan}, semantic global descriptors such as Semantic Scan Context~(SSC)~\cite{li2021ssc} and OverlapNet~\cite{chen2021overlapnet}, as well as semantic graph variants (the family to which our method belongs), such as SGPR~\cite{kong2020semantic} and GOSMatch~\cite{zhu2020gosmatch}.
The authors of GOSMatch do not provide details on the exact Euclidean clustering procedure used to obtain their graph descriptor's vertices, so we assume the resulting number of vertices is similar to ours.
Some of these methods do not yield poses, while others only produce a course estimate which would require further refinement that relies on storing denser point cloud representations.
We observe that the size of our graph representation is comparable to that of global-descriptor-based methods; yet, it also provides an accurate 6-\gls{dof} pose estimate (see \cref{sec:exp:poseest}).
Compared to GOSMatch, which also estimates relative position between scans, we see a notable improvement in memory footprint of more than \num{25} times.
With respect to the raw laser data, we reduce an average of \num{123371} 3D points per scan to only $\num{105}\times\num{7}$, thus reducing memory consumption from \SI{1446}{\kilo\byte} to \SI{2.87}{\kilo\byte}.

% \subsection{Training Details and Hyperparameters}
% \label{sec:exp:train}
% 
% To train the PointNet encoder as described in~\cref{sec:method:featureextraction}, we precompute the semantic graphs for \texttt{Sequence 01}, which is not part of the evaluation set. This results in a total of \num{57396} clusters for the classes of choice, each transformed into its centroid's frame of reference. Cluster sizes vary primarily between \num{20} and \num{200} points in the local pointcloud, with very few in the range of several thousand. Note that batch training is critical for SimCLR since each sample in a batch is contrasted with the remaining ones. Thus, we fix the cluster size $|O_i| = 60$, zero-pad smaller clusters, and use farthest point sampling on the rest. During the augmentation step, we apply random rotation and scaling, and add small perturbations to all the unpadded points. We use a batch size $B = 1024$ and set PointNet's global feature size $D$ to \num{32}, since we want our graph descriptors to be compact. The projection function is a small \acrshort{mlp} with one hidden layer, a leaky ReLU activation function and \num{128} output dimension. The network is trained using the normalised temperature-scaled cross entropy loss (NT-Xent)~\cite{chen2020simple}. \gp{Probably unnecessary to talk about Adam optimizer, learning rate etc.} During inference, no augmentation is performed, but the clusters are still padded or downsampled to \num{60} points.

\subsection{Place Recognition}
\label{sec:exp:placerecognition}

\begin{table*}[t]
\renewcommand{\arraystretch}{1.2}
\centering
\footnotesize
\caption{Aggregates of precision-recall performance across six SemanticKITTI sequences~\cite{behley2019semantickitti}. If based on semantics, methods from the upper group use predicted segmentation (e.g. SSC-RN uses RangeNet++~\cite{milioto2019rangenet}). The lower group represent an upper limit on performance by using ground-truth segmentations. Our method is only outperformed in \texttt{Sequence 08}, which we prove in~\cref{tab:rmse_seq_08} is due to poor segmentation rather than the underlying localisation method -- this is also clear from the best upper bound on performance demonstrated by BoxGraph (BG-SK). \label{tab:fscores}}
\begin{threeparttable}
\begin{tabular}{@{}lccc|ccc|ccc|ccc|ccc|ccc@{}}
\centering
Sequence $\rightarrow$ & \multicolumn{3}{c}{\texttt{00}} & \multicolumn{3}{c}{\texttt{02}} & \multicolumn{3}{c}{\texttt{05}} & \multicolumn{3}{c}{\texttt{06}} & \multicolumn{3}{c}{\texttt{07}} & \multicolumn{3}{c}{\texttt{08}} \\
\toprule
Method & $F_1$ & $F_{0.5}$ & $F_2$ & $F_1$ & $F_{0.5}$ & $F_2$ & $F_1$ & $F_{0.5}$ & $F_2$ & $F_1$ & $F_{0.5}$ & $F_2$ & $F_1$ & $F_{0.5}$ & $F_2$ & $F_1$ & $F_{0.5}$ & $F_2$ \\ \midrule
PointNetVLAD~\cite{uy2018pointnetvlad} & .78 & .82 & .61 & .73 & .80 & .56 & .54 & .63 & .42 & .85 & .87 & .68 & .63 & .72 & .50 & .04 & .03 & .05\\
M2DP~\cite{he2016m2dp} & .71 & .78 & .55 & .72 & .78 & .56 & .60 & .71 & .47 & .79 & .85 & .59 & .56 & .69 & .40 & .07 & .13 & .07\\
IRIS~\cite{wang2021iris} & .67 & .70 & .57 & .76 & .82 & .60 & .77 & .83 & .58 & .91 & .90 & .74 & .63 & .69 & .50 & .48 & .59 & .38\\
ScanContext~\cite{kim2018scan} & .75 & .78 & .62 & .78 & .82 & .63 & .89 & .90 & .72 & .97 & .94 & .78 & .66 & .66 & .54 & .61 & .65 & .52\\
ISC~\cite{wang2020isc} & .66 & .73 & .57 & .71 & .78 & .59 & .77 & .83 & .60 & .84 & .88 & .65 & .64 & .69 & .49 & .41 & .49 & .35\\
SSC-RN~\cite{li2021ssc} & .94 & .92 & .76 & .89 & .89 & .72 & .94 & .93 & .75 & .99 & .95 & .79 & .87 & .89 & .71 & \textbf{.88} & \textbf{.89} & \textbf{.71}\\
OverlapNet~\cite{chen2021overlapnet} & .87 & .87 & .71 & .83 & .86 & .64 & .92 & .91 & .74 & .93 & .91 & .75 & .82 & .80 & .69 & .37 & .43 & .32\\
SGPR~\cite{kong2020semantic} & .82 & .81 & .69 & .75 & .73 & .65 & .75 & .78 & .61 & .65 & .65 & .59 & .87 & .88 & .69 & .75 & .71 & .65\\
\textit{BG-RN (Ours)} & \textbf{.99} & \textbf{.96} & \textbf{.79} & \textbf{.97} & \textbf{.94} & \textbf{.77} & \textbf{.96} & \textbf{.94} & \textbf{.76} & \textbf{1.00} & \textbf{.96} & \textbf{.80} & \textbf{1.00} & \textbf{.96} & \textbf{.80} & .79 & .85 & .59\\
\midrule
SSC-SK~\cite{li2021ssc} & .95 & .93 & .77 & .89 & .88 & .73 & .95 & .93 & .76 & .98 & .95 & .79 & .88 & .89 & .72 & .94 & .93 & .74\\
\textit{BG-SK (Ours)} & \textbf{1.00} & \textbf{.96} & \textbf{.80} & \textbf{.96} & \textbf{.94} & \textbf{.77} & \textbf{.97} & \textbf{.95} & \textbf{.77} & \textbf{1.00} & \textbf{.96} & \textbf{.80} & \textbf{1.00} & \textbf{.96} & \textbf{.80} & \textbf{.96} & \textbf{.94} & \textbf{.76}\\
\bottomrule
\end{tabular}
\end{threeparttable}
\vspace{-10pt}
\end{table*}
\begin{table*}[b]
\renewcommand{\arraystretch}{1.2}
\centering
\footnotesize
\caption{
Aggregates of precision-recall performance across six SemanticKITTI sequences, please refer to~\cref{tab:fscores} for instructions on how to read these results.
In this, $R_1$ refers to recall at \SI{100}{\percent} precision. \label{tab:ap_metrics}}
\begin{threeparttable}
\begin{tabular}{@{}lccc|ccc|ccc|ccc|ccc|ccc@{}}
\centering
% \toprule
Sequence $\rightarrow$ & \multicolumn{3}{c}{\texttt{00}}  & \multicolumn{3}{c}{\texttt{02}}  & \multicolumn{3}{c}{\texttt{05}}  & \multicolumn{3}{c}{\texttt{06}}  & \multicolumn{3}{c}{\texttt{07}}  & \multicolumn{3}{c}{\texttt{08}}  \\
\toprule
Method & $R_1$ & $AP$ & $EP$ & $R_1$ & $AP$ & $EP$ & $R_1$ & $AP$ & $EP$ & $R_1$ & $AP$ & $EP$ & $R_1$ & $AP$ & $EP$ & $R_1$ & $AP$ & $EP$ \\ \midrule
PointNetVLAD~\cite{uy2018pointnetvlad} & .28 & .81 & .64 & .38 & .73 & .69 & .07 & .54 & .54 & .53 & .89 & .77 & .18 & .67 & .59 & .50 & .01 & .25\\
M2DP~\cite{he2016m2dp} & .23 & .73 & .62 & .21 & .71 & .60 & .22 & .62 & .61 & .36 & .77 & .68 & .17 & .49 & .59 & .50 & .04 & .25\\
IRIS~\cite{wang2021iris} & .25 & .74 & .63 & .33 & .79 & .67 & .49 & .80 & .75 & .58 & .97 & .79 & .30 & .66 & .65 & .12 & .48 & .56\\
ScanContext~\cite{kim2018scan} & .22 & .83 & .61 & .26 & .82 & .63 & .59 & .95 & .80 & .85 & .99 & .92 & .11 & .66 & .55 & .14 & .64 & .57\\
ISC~\cite{wang2020isc} & .25 & .74 & .63 & .22 & .76 & .61 & .45 & .83 & .73 & .63 & .88 & .82 & .28 & .66 & .64 & .09 & .40 & .54\\
SSC-RN~\cite{li2021ssc} & .65 & .99 & .83 & .49 & .95 & .74 & .80 & \textbf{.98} & .90 & .95 & .99 & .97 & .54 & .95 & .77 & .46 & \textbf{.95} & .73\\
OverlapNet~\cite{chen2021overlapnet} & .11 & .93 & .56 & .28 & .84 & .64 & .59 & .96 & .80 & .49 & .98 & .74 & .17 & .89 & .59 & .00 & .29 & .17\\
SGPR~\cite{kong2020semantic} & .00 & .88 & .50 & .50 & .74 & .75 & .06 & .80 & .53 & .05 & .68 & .48 & .44 & .91 & .72 & .04 & .78 & .52\\
\textit{BG-RN (Ours)} & \textbf{.98} & \textbf{1.00} & \textbf{.99} & \textbf{.86} & \textbf{.98} & \textbf{.93} & \textbf{.91} & .96 & \textbf{.95} & \textbf{.99} & \textbf{1.00} & \textbf{.99} & \textbf{.99} & \textbf{1.00} & \textbf{1.00} & \textbf{.58} & .76 & \textbf{.79}\\
\midrule
SSC-SK~\cite{li2021ssc} & .70 & .99 & .85 & .50 & .96 & .75 & .81 & \textbf{.99} & .90 & .94 & \textbf{1.00} & .97 & .61 & .96 & .80 & .86 & \textbf{.97 }& \textbf{.93}\\
\textit{BG-SK (Ours)} & \textbf{.98} & \textbf{1.00} & \textbf{.99} & \textbf{.78} & \textbf{.98} & \textbf{.89} & \textbf{.94} & .97 & \textbf{.97} & \textbf{.99} & \textbf{1.00} & \textbf{1.00} & \textbf{1.00} & \textbf{1.00} & \textbf{1.00} & \textbf{.87} & .96 & \textbf{.93}\\
\bottomrule
\end{tabular}
\end{threeparttable}
\end{table*}
\begin{figure*}[t]
\usetikzlibrary{spy}

\centering
\newenvironment{customlegend}[1][]{%
    \begingroup
    \csname pgfplots@init@cleared@structures\endcsname
    \pgfplotsset{#1}%
}{%
    \csname pgfplots@createlegend\endcsname
    \endgroup
}%
\def\addlegendimage{\csname pgfplots@addlegendimage\endcsname}
\begin{tikzpicture}
\begin{customlegend}[legend columns=6,legend style={nodes={scale=0.7, transform shape},align=left,draw=none,column sep=2ex},
legend entries={
{PointNetVLAD~\cite{uy2018pointnetvlad}},
{M2DP~\cite{he2016m2dp}},
{IRIS~\cite{wang2021iris}},
{ScanContext~\cite{kim2018scan}},
{ISC~\cite{wang2020isc}},
{OverlapNet~\cite{chen2021overlapnet}},
{SGPR~\cite{kong2020semantic}},
{SSC-RN~\cite{li2021ssc}},
{SSC-SK~\cite{li2021ssc}},
{\textit{BG-RN (Ours)}},
{\textit{BG-SK (Ours)}},
}]
\addlegendimage{mark=None,pink,solid,thick}
\addlegendimage{mark=none,magenta,solid,thick}   
\addlegendimage{mark=none,black,solid,thick}   
\addlegendimage{mark=none,brown,solid,thick}   
\addlegendimage{mark=none,cyan,solid,thick}   
\addlegendimage{mark=none,purple,solid,thick}   
\addlegendimage{mark=none,orange,solid,thick}   
\addlegendimage{mark=none,green,solid,thick}   
\addlegendimage{mark=none,blue,solid,thick}   
\addlegendimage{mark=none,red,solid,thick}   
\addlegendimage{mark=none,teal,solid,thick}
\end{customlegend}
\end{tikzpicture}
\subfloat[Sequence \texttt{00}\label{fig:seq00}]{
\begin{tikzpicture} [spy using outlines={circle, magnification=3, size=1cm, connect spies}]
\begin{axis}[
width=6cm,
height=4cm,
xmin=0, xmax=1,
ymin=0, ymax=1,
grid=major
]
\addplot+[mark=none,pink,solid,thick] table {data/00/PointNetVlad.txt};
\addplot+[mark=none,magenta,solid,thick] table {data/00/M2DP.txt};
\addplot+[mark=none,black,solid,thick] table {data/00/IRIS.txt};
\addplot+[mark=none,brown,solid,thick] table {data/00/SC.txt};
\addplot+[mark=none,cyan,solid,thick] table {data/00/ISC.txt};
\addplot+[mark=none,purple,solid,thick] table {data/00/overlap.txt};
\addplot+[mark=none,orange,solid,thick] table {data/00/SGPR.txt};
\addplot+[mark=none,green,solid,thick] table {data/00/SSC_rn.txt};
\addplot+[mark=none,blue,solid,thick] table {data/00/SSC.txt};
\addplot+[mark=none,red,solid,thick] table {data/00/box-graph-darknet-knn.txt};
\addplot+[mark=none,teal,solid,thick] table {data/00/box-graph.txt};

\begin{scope}
    \spy[green!70!black,size=2cm] on (4.2,2.2) in node [fill=white] at (1.2,1.2);
\end{scope}

\end{axis}
\end{tikzpicture}
}
\subfloat[Sequence \texttt{02}\label{fig:seq02}]{
\begin{tikzpicture}
\begin{axis}[
width=6cm,
height=4cm,
xmin=0, xmax=1,
ymin=0, ymax=1,
grid=major
]
\addplot+[mark=none,pink,solid,thick] table {data/02/PointNetVlad.txt};
\addplot+[mark=none,magenta,solid,thick] table {data/02/M2DP.txt};
\addplot+[mark=none,black,solid,thick] table {data/02/IRIS.txt};
\addplot+[mark=none,brown,solid,thick] table {data/02/SC.txt};
\addplot+[mark=none,cyan,solid,thick] table {data/02/ISC.txt};
\addplot+[mark=none,purple,solid,thick] table {data/02/overlap.txt};
\addplot+[mark=none,orange,solid,thick] table {data/02/SGPR.txt};
\addplot+[mark=none,green,solid,thick] table {data/02/SSC_rn.txt};
\addplot+[mark=none,blue,solid,thick] table {data/02/SSC.txt};
\addplot+[mark=none,red,solid,thick] table {data/02/box-graph-darknet-knn.txt};
\addplot+[mark=none,teal,solid,thick] table {data/02/box-graph.txt};
\end{axis}
\end{tikzpicture}
}
\subfloat[Sequence \texttt{05}\label{fig:seq05}]{
\begin{tikzpicture} [spy using outlines={circle, magnification=3, size=1cm, connect spies}]
\begin{axis}[
width=6cm,
height=4cm,
xmin=0, xmax=1,
ymin=0, ymax=1,
grid=major
]
\addplot+[mark=none,pink,solid,thick] table {data/05/PointNetVlad.txt};
\addplot+[mark=none,magenta,solid,thick] table {data/05/M2DP.txt};
\addplot+[mark=none,black,solid,thick] table {data/05/IRIS.txt};
\addplot+[mark=none,brown,solid,thick] table {data/05/SC.txt};
\addplot+[mark=none,cyan,solid,thick] table {data/05/ISC.txt};
\addplot+[mark=none,purple,solid,thick] table {data/05/overlap.txt};
\addplot+[mark=none,orange,solid,thick] table {data/05/SGPR.txt};
\addplot+[mark=none,green,solid,thick] table {data/05/SSC_rn.txt};
\addplot+[mark=none,blue,solid,thick] table {data/05/SSC.txt};
\addplot+[mark=none,red,solid,thick] table {data/05/box-graph-darknet-knn.txt};
\addplot+[mark=none,teal,solid,thick] table {data/05/box-graph.txt};

\begin{scope}
    \spy[green!70!black,size=2cm] on (4.2,2.2) in node [fill=white] at (1.2,1.2);
\end{scope}

\end{axis}
\end{tikzpicture}
}

\subfloat[Sequence \texttt{06}\label{fig:seq06}]{
\begin{tikzpicture} [spy using outlines={circle, magnification=3, size=1cm, connect spies}]
\begin{axis}[
width=6cm,
height=4cm,
xmin=0, xmax=1,
ymin=0, ymax=1,
grid=major
]
\addplot+[mark=none,pink,solid,thick] table {data/06/PointNetVlad.txt};
\addplot+[mark=none,magenta,solid,thick] table {data/06/M2DP.txt};
\addplot+[mark=none,black,solid,thick] table {data/06/IRIS.txt};
\addplot+[mark=none,brown,solid,thick] table {data/06/SC.txt};
\addplot+[mark=none,cyan,solid,thick] table {data/06/ISC.txt};
\addplot+[mark=none,purple,solid,thick] table {data/06/overlap.txt};
\addplot+[mark=none,orange,solid,thick] table {data/06/SGPR.txt};
\addplot+[mark=none,green,solid,thick] table {data/06/SSC_rn.txt};
\addplot+[mark=none,blue,solid,thick] table {data/06/SSC.txt};
\addplot+[mark=none,red,solid,thick] table {data/06/box-graph-darknet-knn.txt};
\addplot+[mark=none,teal,solid,thick] table {data/06/box-graph.txt};

\begin{scope}
    \spy[green!70!black,size=2cm] on (4.2,2.2) in node [fill=white] at (1.2,1.2);
\end{scope}

\end{axis}
\end{tikzpicture}
}
\subfloat[Sequence \texttt{07}\label{fig:seq07}]{
\begin{tikzpicture} [spy using outlines={circle, magnification=3, size=1cm, connect spies}]
\begin{axis}[
width=6cm,
height=4cm,
xmin=0, xmax=1,
ymin=0, ymax=1,
grid=major
]
\addplot+[mark=none,pink,solid,thick] table {data/07/PointNetVlad.txt};
\addplot+[mark=none,magenta,solid,thick] table {data/07/M2DP.txt};
\addplot+[mark=none,black,solid,thick] table {data/07/IRIS.txt};
\addplot+[mark=none,brown,solid,thick] table {data/07/SC.txt};
\addplot+[mark=none,cyan,solid,thick] table {data/07/ISC.txt};
\addplot+[mark=none,purple,solid,thick] table {data/07/overlap.txt};
\addplot+[mark=none,orange,solid,thick] table {data/07/SGPR.txt};
\addplot+[mark=none,green,solid,thick] table {data/07/SSC_rn.txt};
\addplot+[mark=none,blue,solid,thick] table {data/07/SSC.txt};
\addplot+[mark=none,red,solid,thick] table {data/07/box-graph-darknet-knn.txt};
\addplot+[mark=none,teal,solid,thick] table {data/07/box-graph.txt};

\begin{scope}
    \spy[green!70!black,size=2cm] on (4.2,2.2) in node [fill=white] at (1.2,1.2);
\end{scope}

\end{axis}
\end{tikzpicture}
}
\subfloat[Sequence \texttt{08}\label{fig:seq08}]{
\begin{tikzpicture}
\begin{axis}[
width=6cm,
height=4cm,
xmin=0, xmax=1,
ymin=0, ymax=1,
grid=major
]
\addplot+[mark=none,pink,solid,thick] table {data/08/PointNetVlad.txt};
\addplot+[mark=none,magenta,solid,thick] table {data/08/M2DP.txt};
\addplot+[mark=none,black,solid,thick] table {data/08/IRIS.txt};
\addplot+[mark=none,brown,solid,thick] table {data/08/SC.txt};
\addplot+[mark=none,cyan,solid,thick] table {data/08/ISC.txt};
\addplot+[mark=none,purple,solid,thick] table {data/08/overlap.txt};
\addplot+[mark=none,orange,solid,thick] table {data/08/SGPR.txt};
\addplot+[mark=none,green,solid,thick] table {data/08/SSC_rn.txt};
\addplot+[mark=none,blue,solid,thick] table {data/08/SSC.txt};
\addplot+[mark=none,red,solid,thick] table {data/08/box-graph-darknet-knn.txt};
\addplot+[mark=none,teal,solid,thick] table {data/08/box-graph.txt};

\end{axis}
\end{tikzpicture}
}
\caption{
Precision-recall curves across six Semantic KITTI Sequences~\cite{behley2019semantickitti}.
These curves are aggregated in~\cref{tab:fscores,tab:ap_metrics}.
Cluttered results are presented on a zoomed in scope, for better readability.
\label{fig:precision_recal_eval}}
\vspace{-15pt}
\end{figure*}
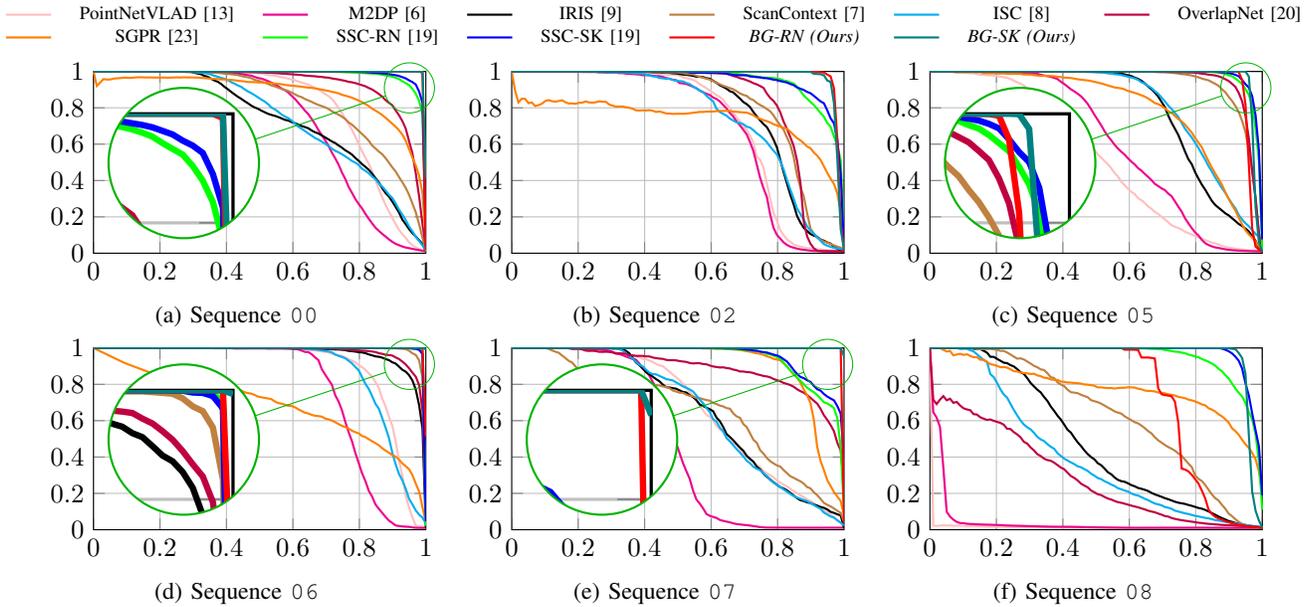

% To assess the place recognition performance of our system, we calculate the similarity scores between each pair of scans in a given sequence, and experiment with different similarity thresholds $\tau$ to determine if the two scans correspond to the same place.
To assess the place recognition performance of our system, we estimate the similarity score $\omega_{s, m}$ (see~\cref{sec:simscore}) between pairs of graphs from each sequence, in turn.
We follow the same setup as in SSC~\cite{li2021ssc} and SGPR~\cite{kong2020semantic}, in which a pair of scans are considered to be a true-positive place recognition if their relative distance is within \SI{3}{\metre}, and a negative if the distance is outside \SI{20}{\metre}.
In particular, we use the benchmarks and evaluation pairs provided by SSC~\cite{li2021ssc} which, for each sequence \texttt{s}, include all $N_s$ positive pairs of scans more than \num{50} frames apart, and randomly sampled $100 \cdot N_s$ negative pairs.
We benchmark our approach against PointNetVLAD~\cite{uy2018pointnetvlad}, M2DP~\cite{he2016m2dp}, LiDAR-Iris~\cite{wang2021iris}, Scan Context~\cite{kim2018scan}, Intensity Scan Context~(ISC)~\cite{wang2020isc}, Semantic Scan Context~(SSC)~\cite{li2021ssc}, OverlapNet~\cite{chen2021overlapnet} and SGPR~\cite{kong2020semantic}.
We compare with SSC both using ground-truth labels from SemanticKITTI (SSC-SK) and predictions from RangeNet++ (SSC-RN).
We show the precision-recall curves and present aggregate metrics, including maximum F-scores, given by
\begin{align*}
F_1=2\frac{PR}{P+R}\\
F_{\beta}=(1+\beta^2)\frac{PR}{\beta^2P+R}
\end{align*}
as well as recall at \SI{100}{\percent} precision ($R_1$), Average Precision ($AP$) and Extended Precision~\cite{ferrarini2020extendedprecision} ($EP$),
$$
AP=\sum_{n}(R_n-R_{n-1})P_n,~EP=\frac{1}{2}(R_1+P_0)
$$
% and Extended Precision~\cite{ferrarini2020extendedprecision}, given by
% \begin{equation}
% EP=\frac{1}{2}(R_1+P_0)
% \end{equation}
where $P_0$ is the precision at \SI{0}{\percent} recall.

% Here,~\cref{fig:precision_recal_eval} shows precision-recall curves,~\cref{tab:fscores,tab:ap_metrics} present aggregates of these curves, and~\cref{tab:rmse_seq_08} presents similar aggregates for Sequence \texttt{08} (see the ablation study below in~\cref{sec:exp:objectablation}).

\cref{fig:precision_recal_eval,tab:fscores,tab:ap_metrics} show that we outperform all other methods in every aggregate and over every sequence.
For example, in \texttt{Sequence 00}, we obtain as much as \num{0.98} $R_1$.
To be clear, this means that the system may be tuned to make \emph{no mistakes} across \SI{98}{\percent} of a large-scale urban autonomous operation, where we only allow it to assume that the single most similar scan is a \textit{bona fide} match.
% Inspecting the precision-recall curves~\cref{fig:precision_recal_eval}, we observe that our method overall has very high recall at \SI{100}{\percent} precision, after which we see a steep drop as recall increases, while other methods have smoother curves.
% This suggests that our descriptor is very descriptive until a certain threshold.
The worst-performing sequence on RangeNet++ predictions is \texttt{Sequence 08}, which contains entirely reverse loop closures.
The recall here is \num{0.58}.
Using ground truth, however, demonstrates comparable results with the other sequences -- \num{0.87} -- suggesting that the lesser performance is due to incorrect predictions by the network (see~\cref{sec:exp:objectablation}).
Contrast this with M2DP and OverlapNet, which are degraded under revisits in the opposite direction.

We obtain a max $F_1$ score of \num{0.951} and $R_1$ of \num{0.884}
for \textit{BG-RN (Ours)}, \textit{averaged across all sequences}.
This is in contrast with \num{0.766} and \num{0.182}, respectively, for SGPR, another semantic-graph-based approach.
The low recall means that SGPR would be prone to false positive loop closures.
Our closest competitor is SSC-RN with average maximum $F_1$ score of \num{0.918} and $R_1$ of \num{0.649}.
Importantly, we tend to outperform SSC-SK (which uses ground truth segmentations), even when we do not use ground truth segmentations (BG-RN).

% For \cref{tab:fscores}, \textit{averaged across all sequences} and excluding SSC-SK and \textit{BG-SK} which use ground truth input segmentations, there is a maximum $F_1$ score of \num{0.951} for \textit{BG-RN (Ours)}, compared to \num{0.703} for IRIS, \num{0.670} for ISC, \num{0.575} for M2DP, \num{0.790} for OverlapNet, \num{0.594} for PointNetVLAD, \num{0.777} for ScanContext, and \num{0.766} for SGPR.
% Our closest competitor is SSC with \num{0.918}
% Also, for \cref{tab:ap_metrics}, \textit{averaged across all sequences} there is a recall at \SI{100}{\percent} precision of \num{0.884} for \textit{BG-RN (Ours)}, compared to \num{0.348} for IRIS, \num{0.321} for ISC, \num{0.282} for M2DP, \num{0.273} for OverlapNet, \num{0.325} for PointNetVLAD, \num{0.362} for ScanContext, and \num{0.182} for SGPR.
% Our closest competitor is SSC with \num{0.649}

\subsection{Pose Estimation}
\label{sec:exp:poseest}

\begin{table*}[t]
\renewcommand{\arraystretch}{1.2}
\centering
\footnotesize
\caption{
\acrshort{rte} (top, values in \SI{}{\metre}) and \acrshort{rre} (bottom, in \SI{}{\deg}) pose estimation results, showing median ($q_2$) as well as lower and upper quartiles ($q_1$ and $q_3$) per SemanticKITTI sequence.
Bold shows best performing, while underlined shows second-best performing.
\textit{BG-RN (Ours)} from the upper groups uses predicted segmentation (FPFH and FCGF are not based on semantics).
The lower groups represent an upper limit on performance by using ground-truth segmentations.
Even though SSC uses ground truth (i.e. SSC-SK), our system lacking ground truth (\textit{BG-RN}), consistently outperforms it by a large margin.
It is important to note that the upper limit on performance for our system \textit{BG-SK} in fact outperforms the learned FCGF.
% There is a mixed result for Sequence~\texttt{08}, although in the presence of beter segmentation our system outperforms all others, which is confirmed in the ablation study of~\cref{tab:rmse_seq_08}.
\label{tab:rte_rre}
}

\begin{subtable}[h]{\textwidth}
\centering
\begin{threeparttable}
\centering
\begin{tabular}{@{}lccc|ccc|ccc|ccc|ccc|ccc@{}}
Sequence $\rightarrow$ & \multicolumn{3}{c}{\texttt{00}}  & \multicolumn{3}{c}{\texttt{02}}  & \multicolumn{3}{c}{\texttt{05}}  & \multicolumn{3}{c}{\texttt{06}}  & \multicolumn{3}{c}{\texttt{07}}  & \multicolumn{3}{c}{\texttt{08}}  \\
\toprule
Method & $q_2$ & $q_1$ & $q_3$ & $q_2$ & $q_1$ & $q_3$ & $q_2$ & $q_1$ & $q_3$ & $q_2$ & $q_1$ & $q_3$ & $q_2$ & $q_1$ & $q_3$ & $q_2$ & $q_1$ & $q_3$ \\ \midrule
FPFH~\cite{rusu2009fpfh}  & .13 & .09 & .19 & .14 & .09 & .22 & .12 & .08 & .19 & .17 & .11 & .26 & .11 & .07 & .17 & \underline{.16} & .10 & \underline{.25}\\
FCGF~\cite{choy2019fully}  & \textbf{.08} & \textbf{.05} & \textbf{.11} & \textbf{.09} & \textbf{.06} & \textbf{.15} & \textbf{.07} & \textbf{.05} & \textbf{.10} & \textbf{.06} & \textbf{.05} & \textbf{.09} & \underline{.08} & \textbf{.05} & \textbf{.11} & \textbf{.08} & \textbf{.05} & \textbf{.12}\\
\textit{BG-RN (Ours)}  & \underline{.10} & \underline{.06} & \underline{.17} & \underline{.11} & \underline{.07} & \underline{.19} & \underline{.09} & \textbf{.05} & \underline{.15} & \underline{.11} & \underline{.07} & \underline{.19} & \textbf{.07} & \underline{.05} & \underline{.12} & \underline{.16} & \underline{.09} & .48\\
\midrule
SSC-SK~\cite{li2021ssc}  & .17 & .09 & .34 & .21 & .10 & .47 & .15 & .07 & .33 & .19 & .11 & .36 & .17 & .07 & .44 & .18 & .09 & .36\\
\textit{BG-SK (Ours)}  & \textbf{.08} & \textbf{.05} & \textbf{.12} & \textbf{.08} & \textbf{.05} & \textbf{.13} & \textbf{.07} & \textbf{.04} & \textbf{.10} & \textbf{.06} & \textbf{.04} & \textbf{.09} & \textbf{.06} & \textbf{.04} & \textbf{.10} & \textbf{.08} & \textbf{.05} & \textbf{.12}\\
\bottomrule
\end{tabular}
\end{threeparttable}
\end{subtable}

\vspace{5pt}

\begin{subtable}[h]{\textwidth}
\centering
\begin{threeparttable}
\centering
\label{tab:rot}
\begin{tabular}{@{}lccc|ccc|ccc|ccc|ccc|ccc@{}}
\centering
Sequence $\rightarrow$ & \multicolumn{3}{c}{\texttt{00}}  & \multicolumn{3}{c}{\texttt{02}}  & \multicolumn{3}{c}{\texttt{05}}  & \multicolumn{3}{c}{\texttt{06}}  & \multicolumn{3}{c}{\texttt{07}}  & \multicolumn{3}{c}{\texttt{08}}  \\
\toprule
Method & $q_2$ & $q_1$ & $q_3$ & $q_2$ & $q_1$ & $q_3$ & $q_2$ & $q_1$ & $q_3$ & $q_2$ & $q_1$ & $q_3$ & $q_2$ & $q_1$ & $q_3$ & $q_2$ & $q_1$ & $q_3$ \\ \midrule
FPFH~\cite{rusu2009fpfh}  & \underline{.35} & .22 & \underline{.56} & .38 & .24 & \underline{.64} & .32 & .19 & .54 & .39 & .23 & .64 & .28 & .16 & \underline{.47} & \underline{.46} & .27 & \underline{.84}\\
FCGF~\cite{choy2019fully}  & \textbf{.18} & \textbf{.11} & \textbf{.29} & \textbf{.21} & \textbf{.13} & \textbf{.38} & \textbf{.14} & \textbf{.08} & \textbf{.24} & \textbf{.11} & \textbf{.07} & \textbf{.16} & \textbf{.16} & \textbf{.09} & \textbf{.28} & \textbf{.17} & \textbf{.10} & \textbf{.30}\\
\textit{BG-RN (Ours)}  & .36 & \underline{.19} & .68 & \underline{.34} & \underline{.18} & .70 & \underline{.25} & \underline{.10} & \underline{.52} & \underline{.26} & \underline{.13} & \underline{.51} & \underline{.22} & \underline{.08} & \underline{.47} & .49 & \underline{.22} & 1.35\\
\midrule
SSC-SK~\cite{li2021ssc}  & .56 & .32 & .95 & .46 & .26 & .84 & .50 & .21 & .86 & .40 & .19 & .73 & .44 & .14 & .90 & .58 & .30 & 1.05\\
\textit{BG-SK (Ours)}  & \textbf{.29} & \textbf{.16} & \textbf{.50} & \textbf{.26} & \textbf{.14} & \textbf{.49} & \textbf{.21} & \textbf{.10} & \textbf{.37} & \textbf{.16} & \textbf{.10} & \textbf{.24} & \textbf{.20} & \textbf{.08} & \textbf{.39} & \textbf{.27} & \textbf{.15} & \textbf{.50}\\
\bottomrule
\end{tabular}
\end{threeparttable}
\end{subtable}

\vspace{-10pt}
\end{table*}
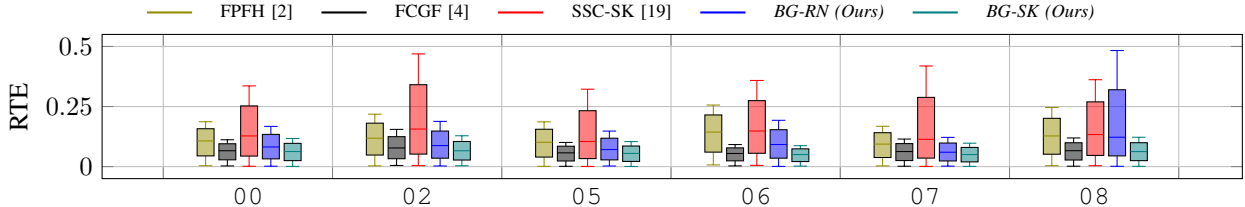
\begin{figure*}[b]
\centering
\vspace{-5pt}
\newenvironment{customlegend}[1][]{%
    \begingroup
    \csname pgfplots@init@cleared@structures\endcsname
    \pgfplotsset{#1}%
}{%
    \csname pgfplots@createlegend\endcsname
    \endgroup
}%
\def\addlegendimage{\csname pgfplots@addlegendimage\endcsname}
\begin{tikzpicture}
\begin{customlegend}[legend columns=6,legend style={nodes={scale=0.7, transform shape},align=left,draw=none,column sep=2ex},
legend entries={
{FPFH~\cite{rusu2009fpfh}},
{FCGF~\cite{choy2019fully}},
{SSC-SK~\cite{li2021ssc}},
{\textit{BG-RN (Ours)}},
{\textit{BG-SK (Ours)}},
}]
\addlegendimage{mark=None,olive,solid,thick}
\addlegendimage{mark=none,black,solid,thick}
\addlegendimage{mark=none,red,solid,thick}
\addlegendimage{mark=none,blue,solid,thick}   
\addlegendimage{mark=none,teal,solid,thick}   
\end{customlegend}
\end{tikzpicture}

\begin{tikzpicture}
\begin{axis}[
every axis plot/.append style={fill,fill opacity=0.5},
boxplot/draw direction=y,
ylabel={RTE},
height=3.5cm,
width=16cm,
cycle list={{olive},{black},{red},{blue},{teal}},
boxplot={
draw position={1/4+1/5*\numplotsofactualtype - 1/14*mod(\numplotsofactualtype, 5)},
box extend=0.1,
every box/.style={draw=black},
},
x=2.25cm,
xtick={0,1,2,...,25},
ytick={0,0.25,...,1},
x tick label as interval,
ymin=0,ymax=0.55,
xticklabels={%
{\texttt{00}},%
{\texttt{02}},%
{\texttt{05}},%
{\texttt{06}},%
{\texttt{07}},%
{\texttt{08}},%
},
x tick label style={
text width=2.5cm,
align=center,
grid=major,
ymin=-0.05
},
]

%----------------------%
%boxtop whiskertop median boxbottom whiskerbottom
%----------------------%
\addplot table[row sep=\\,y index=0] { data\\0.18677728756041495\\30.406278120389807\\0.12848706316840977\\0.08546960570328435\\0.0036522690303432332\\};
\addplot table[row sep=\\,y index=0] { data\\0.1117703618641368\\15.502347922411994\\0.0784530840783294\\0.0535861717341079\\0.002838672517060327\\};
\addplot table[row sep=\\,y index=0] { data\\0.33590469422984515\\6.312016194387883\\0.16932959802946662\\0.08614275186602409\\0.0014976508472826426\\};
\addplot table[row sep=\\,y index=0] { data\\0.16738344205530997\\69.0537614609103\\0.10064299345565414\\0.06284012101622527\\0.0016491734858686922\\};
\addplot table[row sep=\\,y index=0] { data\\0.11704385428789396\\43.54075236684562\\0.07652500706672315\\0.04895347960678985\\0.0008120289664719457\\};
\addplot table[row sep=\\,y index=0] { data\\0.21814999042649957\\49.41394983164806\\0.14349743459811962\\0.09310387788218635\\0.0031445730072079707\\};
\addplot table[row sep=\\,y index=0] { data\\0.1549113056514686\\31.0070906676014\\0.09435262076498102\\0.060853795149790446\\0.00475392536841793\\};
\addplot table[row sep=\\,y index=0] { data\\0.4690598681765006\\10.11029233073306\\0.21257964343871646\\0.09993910490311951\\0.00409856228646831\\};
\addplot table[row sep=\\,y index=0] { data\\0.18824119321696847\\64.457454731976\\0.10773888029408629\\0.06732743580695448\\0.0027254759831588414\\};
\addplot table[row sep=\\,y index=0] { data\\0.12837994570353486\\73.18357974619573\\0.0806935592248795\\0.051192225230630134\\0.003583307969419128\\};
\addplot table[row sep=\\,y index=0] { data\\0.1863067264546701\\43.5348760715159\\0.12459832534522484\\0.07809656407221767\\0.0011626657358558322\\};
\addplot table[row sep=\\,y index=0] { data\\0.10060841581989913\\3.1046922783910826\\0.06924584814006471\\0.04539389911457865\\0.0012557647506866588\\};
\addplot table[row sep=\\,y index=0] { data\\0.3221977677050891\\6.395642876337833\\0.14304369690198018\\0.0659235485810706\\0.0007018865069168313\\};
\addplot table[row sep=\\,y index=0] { data\\0.14784058654508675\\49.706275478942565\\0.08837795852601091\\0.053575161448718496\\0.002436821474952785\\};
\addplot table[row sep=\\,y index=0] { data\\0.1041683400068811\\44.08301101393845\\0.06727261148958065\\0.04257224508547496\\0.0006014310294603959\\};
\addplot table[row sep=\\,y index=0] { data\\0.2561758594891381\\36.2129073555592\\0.1738325302678388\\0.11348430957324629\\0.007017696767441172\\};
\addplot table[row sep=\\,y index=0] { data\\0.09183322514171108\\2.8258408709629803\\0.06426186450596931\\0.044991704134729985\\0.0029825053395364626\\};
\addplot table[row sep=\\,y index=0] { data\\0.35847956954323656\\6.028561222077252\\0.1904978329511693\\0.10573091373572907\\0.0051973405379662175\\};
\addplot table[row sep=\\,y index=0] { data\\0.19272772617667525\\41.67647693080877\\0.11439957485103629\\0.06939528329266004\\0.0008732426603353506\\};
\addplot table[row sep=\\,y index=0] { data\\0.08716833184931137\\1.0213004439743043\\0.06067759516633684\\0.03989098548793803\\0.0020564574429327493\\};
\addplot table[row sep=\\,y index=0] { data\\0.1678237741381001\\24.96160142275155\\0.11443574110602574\\0.07284694895776493\\0.0029077496242385288\\};
\addplot table[row sep=\\,y index=0] { data\\0.11467060595113399\\5.386691019890058\\0.07684300975518008\\0.04865233189611918\\0.001556269148071501\\};
\addplot table[row sep=\\,y index=0] { data\\0.41874448664568603\\5.693301055681772\\0.15740693767000608\\0.06978604856531416\\0.0009643452864565172\\};
\addplot table[row sep=\\,y index=0] { data\\0.1216509153781521\\39.811922576818176\\0.07451393932835536\\0.045165067248741635\\0.0019491721331017344\\};
\addplot table[row sep=\\,y index=0] { data\\0.09750536688761055\\28.003364281713278\\0.062110689341222616\\0.03840976614341598\\0.0010513040794093518\\};
\addplot table[row sep=\\,y index=0] { data\\0.24585648973470642\\42.8347667730638\\0.15601893701697317\\0.09923530830618378\\0.003760982328847073\\};
\addplot table[row sep=\\,y index=0] { data\\0.1192397252411452\\30.206457356402506\\0.07942830609399001\\0.053066537132354305\\0.0017312143387905273\\};
\addplot table[row sep=\\,y index=0] { data\\0.3615759792973212\\9.892290998264167\\0.17745233353104892\\0.08941637115936016\\0.0037634504060991844\\};
\addplot table[row sep=\\,y index=0] { data\\0.48310087974097543\\69.90682580710553\\0.1565876204536757\\0.08789221068900438\\0.001275281035709686\\};
\addplot table[row sep=\\,y index=0] { data\\0.12218961263592386\\40.52780739123473\\0.07682412939003416\\0.04836421188852162\\0.001242664272864636\\};

\end{axis}

\end{tikzpicture}
\caption{
Quartile ($q_1$, $q_2$, and $q_3$) and full-range ($\operatorname{min}/\operatorname{max}$) RTE statistics for metric pose estimation across six SemanticKITTI sequences~\cite{behley2019semantickitti}.
Note that the results for SSC-SK (red) are generated using \emph{ground-truth} semantic segmentations, while our method (\textit{BG-RN}, blue) uses predictions.
We nevertheless consistently outperform SSC as well as FPFH (orange), except in \texttt{Sequence 08}.
~\cref{tab:rmse_seq_08} shows this is due to the poor quality of input predictions.
% The ablation study in~\cref{tab:rmse_seq_08} examines this sequence further, showing with better predictions that our method was limited in this sequence by poor segmentation and not by the localisation method proposed.
It is also important to note that with ideal segmentations (\textit{BG-SK}), our method is capable of outperfoming even FCGF~\cite{choy2019fully} (black), a state-of-the-art learned approach.
\label{fig:mse_boxplot}}
\end{figure*}

Similarly to~\cref{sec:exp:placerecognition}, we evaluate the metric localisation capabilities of our system by estimating the pose $T_{s, m}$ (see~\cref{sec:poserefine}) between pairs of graphs from each sequence database, in turn.
Here we consider all pairs of distinct scans within ground truth distance of \SI{3}{\metre}.
% and report the \gls{rmse} between the estimates and corresponding ground truth transformations.
We found the ground-truth GPS poses of the KITTI odometry dataset to be noisy, particularly for \texttt{Sequence 00}, so we refine them using the \gls{icp} algorithm.
We compare our performance with two global registration approaches based on matching dense local features with \gls{ransac} -- using the classical \gls{fpfh}~\cite{rusu2009fpfh}, and the learned \gls{fcgf}~\cite{choy2019fully}.
% We compare our performance with a classical global registration approach, FPFH \gls{ransac}~\cite{rusu2009fpfh}, and SSC~\cite{li2021ssc} using ground-truth labels.
For \gls{fpfh} and \gls{fcgf}, we downsample input point clouds with a voxel size of \SI{30}{\centi\metre} and we use \gls{fcgf}'s pre-trained model on KITTI with \num{32} feature dimensions.
For a fair comparison, we set the \gls{ransac} iterations for all methods to \num{10000}.
We also compare with SSC~\cite{li2021ssc} using ground-truth labels as a representative of a fellow semantic-based loop-closure approach that also yields an initial pose, but it is worth noting this pose has only 3-\gls{dof}.
We report the \gls{rte} and \gls{rre} following the definitions from \cite{choy2019fully}:
\begin{equation*}
\small{
    RTE = \norm{t_{s,m} - t^*_{s,m}}_2,~RRE = \cos^{-1}(\frac{\text{Tr}\:(R_{s,m}^T R^*_{s,m}) - 1)}{2})
    }
\end{equation*}
% and
% \begin{equation}
%     \textbf{RRE} = \arccos(\frac{\text{Tr}\:(R_{s,m}^T R^*_{s,m}) - 1)}{2})
% \end{equation}
where $R^*$ and $t^*$ are the ground-truth values for the estimated pose and $\text{Tr}(\cdot)$ is the trace operation.

Here,~\cref{fig:mse_boxplot} shows per-sequence \acrshort{rte} error distributions as box-and-whisker plots.
% \gp{\cref{tab:rmse_seq_08} present \gls{rmse} statistics for Sequence \texttt{08} (see the ablation study below in~\cref{sec:exp:objectablation}).}
We can see that our method performs consistently across all sequences, except \texttt{Sequence 08} when using predictions, which we deduce is due to prediction noise (see~\cref{sec:exp:objectablation}).
SSC-SK performs worst, which suggests it would require further refinement when deployed on a live system.
We can conclude that out of all the place recognition approaches in~\cref{sec:exp:placerecognition}, we are the only one that also produces an accurate pose estimate.
% , using predictions, consistently outperforms both FPFH and SSC baselines.
\cref{tab:rte_rre} shows the median as well as lower and upper quartile of RTE and RRE.
% For example, for Sequence \texttt{05}, we achieve a median \gls{rmse} error (with predicted segmentation, not ground truth) of \num{0.12} as opposed to \num{0.16} for FPFH and \num{0.22} for SSC.
% Importantly, SSC was run with ground-truth predictions here, and out method still outperformed it.
Aggregated over \textit{all sequences}, we have median RTE of 
% \SI{0.134}{\metre} for FPFH,
% \SI{0.077}{\metre} for FCGF,
% \SI{0.170}{\metre} for SSC-SK (with ground-truth segmentations),
% \SI{0.104}{\metre} for BG-RN (ours with \textit{predicted} segmentations),
% \SI{0.072}{\metre} for BG-SK (ours with ground-truth segmentations).
\SI{13.4}{\centi\metre} for FPFH,
\SI{7.7}{\centi\metre} for FCGF,
\SI{17}{\centi\metre} for SSC-SK (with ground-truth segmentations),
\SI{10.4}{\centi\metre} for BG-RN (ours with \textit{predicted} segmentations),
\SI{7.2}{\centi\metre} for BG-SK (ours with ground-truth segmentations).
Similarly, median RRE is
$0.361^{\circ}$ for FPFH,
$0.167^{\circ}$ for FCGF,
$0.513^{\circ}$ for SSC-SK,
$0.331^{\circ}$ for BG-RN,
$0.246^{\circ}$ for BG-SK.
Note that \gls{fpfh} and \gls{fcgf} operate on downsampled scans with around \num{20000} points ($\SI{234.38}{\kilo\byte}$), which would still need to be stored in the map.
As mentioned in~\cref{tab:rte_rre}, when using predictions, we are outperforming non-learned geometry-based methods and approaching the performance of learned variants.
Indeed, it would seem that our method, if presented with ideal segmentations, is capable of better performance than FCGF, a descriptor trained explicitly to register LiDAR scans.
% This is discussed further in~\cref{sec:exp:objectablation}.

% On the same sequence from~\cref{tab:rte_rre} we achieve for \gls{rte} \num{0.1} as opposed to \num{0.14} and \num{0.17} for FPFH and SSC (despite being with ground truth) and for \gls{rre} \num{0.1} as opposed to \num{0.14} and \num{0.17}\gadd{check nums}.

\subsection{Effect of segmentation prediction quality}
\label{sec:exp:objectablation}

We investigate the effect of the performance of the prerequisite segmentation network on our localisation capability, where ground-truth segmentation labels provide an upper limit on performance.
This is done both to contribute to the body of evidence on the applicability of various trained networks and to confirm the efficacy of our underlying localisation formulation beyond any variance in segmentation network performance.
We focus on \texttt{Sequence 08}, the \emph{only} sequence in which we saw a considerable drop in performance when using predicted as opposed to ground-truth labels~\cref{fig:precision_recal_eval,fig:mse_boxplot}.

\cref{tab:rmse_seq_08} shows that our method is improved in its place recognition performance by the use of Cylinder3D~\cite{zhou2020cylinder3d} -- a better segmentation network -- achieving for example a maximum $F_1$ score of \num{0.89} (originally \num{0.79}).
Correspondingly, our median RTE improves from \num{0.16} to \num{0.12}.

\begin{table}[!h]
\renewcommand{\arraystretch}{1.2}
\centering
\resizebox{\columnwidth}{!}{\begin{threeparttable}
\caption{
% Ablation study, investigating of the effect on both place recognition and metric pose estimation performance of better segmentation predictions from the input network.
% Shown here are RMSE errors, as opposed to RTE in~\cref{fig:mse_boxplot}.
Effect on both place recognition and metric pose estimation performance of using better segmentation predictions from the input network (RN -- RangeNet++~\cite{milioto2019rangenet}, CYL3D -- Cylinder3D~\cite{zhou2020cylinder3d}, SK -- ground truth).
% \gadd{Move to top of page?}
\label{tab:rmse_seq_08}
}
\begin{tabular}{@{}l|cccccc|ccc@{}}
\centering
& \multicolumn{6}{c|}{Place recognition} & \multicolumn{2}{c}{Pose estimation} \\
Method & $F_1$ & $F_{.5}$ & $F_2$ & $R_{1}$ & $AP$  & $EP$ & $RTE(\SI{}{\metre})$ & $RRE(\SI{}{\circ})$\\
\midrule
% BG-SVD-RN & .73 & .8 & .55 & .33 & .75 & .66 & 7.88 & 1.28 & 16.96\\
BG-RN & .79 & .85 & .59 & .58 & .76 & .79 & .16 & .49\\
BG-CYL3D & \textbf{.89} & \textbf{.91} & \textbf{.68} & \textbf{.72} & \textbf{.88} & \textbf{.86} & \textbf{.12} & \textbf{.40}\\
\midrule
% BG-SVD-SK & .79 & .85 & .59 & .58 & .76 & .79 & 1.73 & .22 & 8.12\\
BG-SK & \textbf{.96} & \textbf{.94} & \textbf{.76} & \textbf{.87} & \textbf{.96 } & \textbf{.94} & \textbf{.08} & \textbf{.27}\\
\bottomrule
\end{tabular}
\end{threeparttable}}
\vspace{-12pt}
\end{table}

This, alongside the evidence in~\cref{tab:fscores,tab:ap_metrics} in which \textit{BG-SK} outperforms \textit{SSC-SK} \emph{across all aggregates} and \emph{all sequences}, further suggests that our upper limit on performance is superior to our close competitor.
Importantly, \textit{BG-RN} is generally more robust even in the face of poorer segmentation predictions for all other sequences in~\cref{tab:fscores,tab:ap_metrics}.

\section{Conclusion}%
\label{sec:conclusion}
%------------------------------------------------------------------

We have presented a system for extremely lightweight and performant LiDAR localisation using a compact but discriminative representation derived from semantic segmentation of raw laser data.
We adapt a semantic graph descriptor to high-fidelity 3D laser scans for both place recognition and rigid-body registration.
We apply the representation to place recognition using 3D bounding boxes as appearance embeddings of vertex entities and extend the method to pose estimation.
We demonstrate state-of-the-art place recognition with competitive metric localisation performance on the KITTI odometry dataset while representing scans with a fractional memory requirement.
Our method, if presented with ideal segmentations, is capable of better performance than systems trained explicitly to estimate precise pose between dense laser scans, and we expect that, as scene understanding theory and practice develop, the work presented in this paper will be exploited heavily in robust localisation with LiDAR.

% We validate the approach on the KITTI odometry dataset and show competitive results -- both in terms of topological and metric localisation -- with contemporary semantic-based methods -- while representing scans with fractional memory requirement.
% \pmnsays{meta point  -why not sell it as a whole package which always includes local features - then the removal of it is some sort of abalation point. }
% Furthermore, we improve both topological and metric localisation performance by including within the framework features describing local point cloud appearance characteristics to be used alongside the semantic properties of the scanned environment.
% Here, we see boosts to performance when considering data at long-ranges from the ego-vehicle and in cases of map decimation.

% Future work will include applying similar techniques to semantic motion estimation tasks, cross-modal multi-sensor setups, and lifelong navigation during extended multi-session scenarios.

%------------------------------------------------------------------
\section*{Acknowledgements}
%------------------------------------------------------------------

Thanks to the Assuring Autonomy International Programme, a partnership between Lloyd’s Register Foundation and the University of York, and EPSRC Programme Grant ``From Sensing to Collaboration'' (EP/V000748/1).

% \gp{As per review:
% 2. Some references are not in their official format.
% i) Pointnet++ was published at NIPS 2017
% ii) Locus was published at ICRA 2021
% iii)  SSC was published at IROS 2021
% iv) Overlapnet was published at AuRo 2021
% v) YOLOv3 was published at CVPR 2018
% }

%------------------------------------------------------------------
\bibliographystyle{IEEEtran}
\bibliography{biblio}
%------------------------------------------------------------------

\end{document}